# QR Code Denoising using Parallel Hopfield Networks

Ishan Bhatnagar and Shubhang Bhatnagar

*Abstract—* We propose a novel algorithm for using Hopfield networks to denoise QR codes. Hopfield networks have mostly been used as a noise tolerant memory or to solve difficult combinatorial problems. One of the major drawbacks in their use in noise tolerant associative memory is their low capacity of storage, scaling only linearly with the number of nodes in the network. A larger capacity therefore requires a larger number of nodes, thereby reducing the speed of convergence of the network in addition to increasing hardware costs for acquiring more precise data to be fed to a larger number of nodes. Our paper proposes a new algorithm to allow the use of several Hopfield networks in parallel thereby increasing the cumulative storage capacity of the system many times as compared to a single Hopfield network. Our algorithm would also be much faster than a larger single Hopfield network with the same total capacity. This enables their use in applications like denoising QR codes, which we have demonstrated in our paper. We then test our network on a large set of QR code images with different types of noise and demonstrate that such a system of Hopfield networks can be used to denoise and recognize QR code's in real time.

## I. Introduction

In this paper we consider the problem of de-noising QR codes which have recently grown popular due to the explosion of mobile applications. QR codes which stand for "Quick response" codes were invented in Japan by Denso Wave in the year 1994, QR bar codes are a type of matrix 2D bar codes which were originally created to track vehicles during their manufacturing process (see Figure 1). They were specifically designed to allow its contents to be decoded at high speed, they have now become the most popular type of matrix 2D bar codes and are easily read by most smartphones as they allow for data capture at high speeds [1].

There are various versions of QR codes available depending on their information capacity [2]. We have used Version 10 which has a size of 57 x 57 for all further research. Currently these QR codes use the "Reed-Solomon" error correction codes which occupy significant space of the QR code itself [3]. Hence, we propose a novel way of denoising these codes without use of error correcting codes. The proposed algorithm uses Hopfield networks as an associative and content addressable memory (CAM) which are then used to denoise the codes. Hopfield and his colleagues have already shown that a symmetric interconnected neural network (a Hopfield neural network) can perform error corrections in associative retrieval [4].

Ishan Bhatnagar is with the Thadomal Shahani Engineering College, Bandra West, Mumbai(e-mail:ishanb98@gmail.com). Shubhang Bhatanagar is with Indian Institute of Technology, Bombay, Powai ,Mumbai(e-mail:160020019@iitb.ac.in)

The Content Addressable memory (CAM) is by definition a memory in which the stored patterns are reconstructed by presenting the network with a partial form of the pattern which in our case are noisy images of QR codes [5]. We also test our model on various forms of noise and how our model deals with these various forms of noisy QR codes. The Hopfield network model is a single layer of neurons which are fully connected and recurrent i.e. every neuron is connected to each other neuron including themselves as shown in Figure 2. In the figure, a system of 3 neurons is taken. In this notation $x_i$ are the input vectors whereas $y_i$ are the output vectors for the given system. The model can be extended for any general system with any *n* number of neurons. Although we have described a recurrent example of Hopfield network sometimes the self-loops are not used such that a unit does not influence itself.

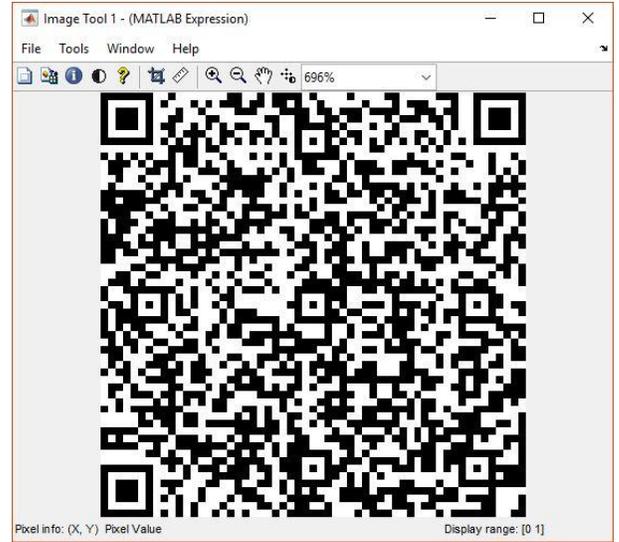

*Figure 1-A sample QR code*

The Hopfield network we use has a set of interconnected neurons whose values are updated asynchronously. The activation values used are bipolar i.e. {+1, -1} and work like a human brain to store and retrieve memory. The general Hopfield model assumes usage of Hebbian learning in which two neurons which fire together simultaneously have their synaptic connections (in our case the weight matrix coefficients) strengthened. We use a modified form of Hebbian learning in our model which is described in the algorithm section of the paper. The term Hebbian learning was used as it was derived from the work of Donald Hebb [6] who hypothesized in 1949 how the neurons are connected and work in the brain.

## II. RELATED WORK

Recently due to the growth of hardware, computational power and simultaneous increase of memory being available has increased the feasibility of the use of Hopfield networks in a real time scenario. Hopfield networks have been recently used as a bidirectional associative memory for semantic neural networks [7]. They have also been used as a content addressable memory to find code words in VLSI hardware structures since code words are essentially stored as a pattern of binary data [8]. We similarly use Hopfield network as an associative and content addressable memory (CAM) to retrieve the correct QR codes given a noisy version of the same codes.

Hopfield networks have also been used to restore various kinds of images recently combined with other algorithms and the basic Hebbian rule of learning [9]. There have also been attempts to increase the capacity of Hopfield networks using various types of genetic algorithms [10]. However, we propose a novel method to increase the capacity of the Hopfield network by distributing the load of one Hopfield network into several parallel Hopfield networks. This also brings about the problem of having multiple energy functions and choosing between them when reconstructing a specific QR code which is discussed in detail in the next section of the paper. Also, several methods have been tried to improve the de-noising of QR codes using traditional error codes and pre-processing techniques [11] but these waste precious space on the QR code which might otherwise have been precious data.

## III. ALGORITHM

Our algorithm for denoising the QR code involves training the network with the QR codes to be recognized. The steps followed for training the network to denoise the QR codes are-

### A. Generation of data

We generate data using an open source QR code generator [12] based on the ZXing library for java. A total of 4,000 QR 57×57 (version 10) QR codes were generated using this QR code generator. The QR codes were generated for a random 200-character string as input. The blank space padding was taken as 0.

### B. Training Algorithm-

We have divided the 4,000 QR codes into 10 sets of 400 each (done randomly). This is required because the number of nodes in the Hopfield network which we use = number of pixels/binary units in the QR code=57*57=3249. The 57×57 QR codes are converted to 1 dimensional 3249×1 vectors before using them for training the network. Also, the maximum capacity of associative memory for such a network, when using the Hebbian rule of learning has an upper bound of $\approx 0.14\,n$ [13]. So, to have a larger capacity, we used the pseudo-inverse rule for training the network [14].

The pseudo-inverse rule involves first training the network using Hebbian rule[9] to obtain a weights matrix W. Then we make the diagonal elements of the matrix equal to zero (to prevent self-connections for extra stability). Then for getting the pseudo-inverse rule's weight matrix, we take the pseudoinverse of the zero diagonal matrix W. Then we again make the diagonal elements of the weight matrix zero (to prevent self-connections for extra stability). For getting $W_{pseudoinverse}$ we do-

$$W_{d,hebbian} = W_{hebbian}\ with\ W_{ii} = 0$$
$$W_{g,pseudoinverse} = pseudoinverse(W_{d,hebbian})$$

$$W_{pseudoinverse} = W_{g,pseudoinverse}\ with\ W_{ii} = 0$$

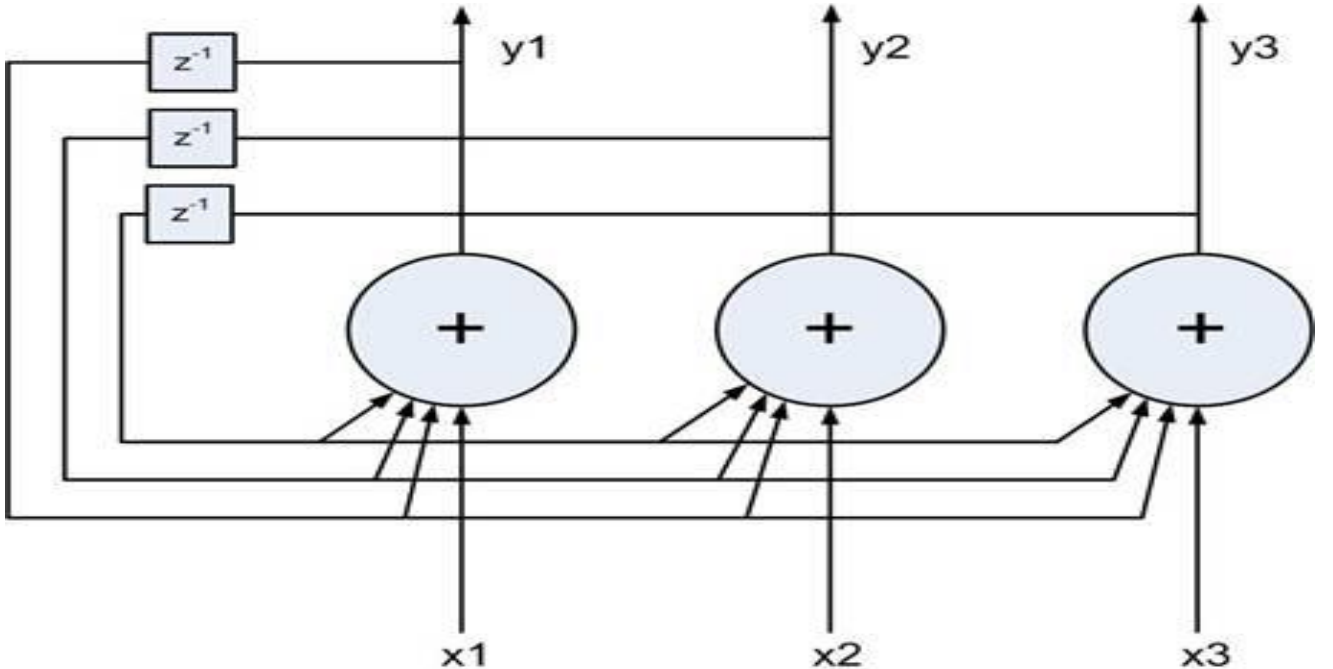

*Figure 2 A general Hopfield network*

But even this capacity(<n) is not large enough for the potentially large number of QR codes which may need to be recognized. So, we distribute the QR codes to be stored in many (10 in number) Hopfield networks randomly, and use our algorithm we describe in the parallel algorithm section to use the appropriate network.

We train a Hopfield network on each of these sets of QR codes individually. Each Hopfield network being trained has 3249 nodes (=number of pixels in the QR code) and as it is a fully connected Recurrent neural network, it results in a weight matrix of size 3249×3249. On training using 10 (in our example) networks on 400 57×57 QR codes each, we get the 10 3249×3249 weight matrices corresponding to the 10 networks. Any QR code is just trained in one of these 10 matrices.

*C. Preparing Noisy QR codes*

We tested our network on QR codes containing 3 different types of noise- gaussian, salt and pepper and huge amounts of localized noise.

a) **Gaussian noise**- We modelled noise in each pixel of the QR code as a gaussian random variable (independent foreach pixel) having zero mean(μ=0) and finite variance (σ2). We empirically selected a variance of 0.3 for the gaussian random variable. We used the imnoise function [15] of MATLAB with the appropriate parameter values selected for making a noisy QR code. After adding the noise, we had to binarize the image (as QR codes are binary images) and the noise added led to multiple pixels getting flipped. Based on the variance which we selected, the fraction of pixels which are flipped are-

$$Noise\ Random\ variable = X$$
$$Let\ Z = \frac{X}{\sqrt{0.3}} = \frac{X}{0.54}$$

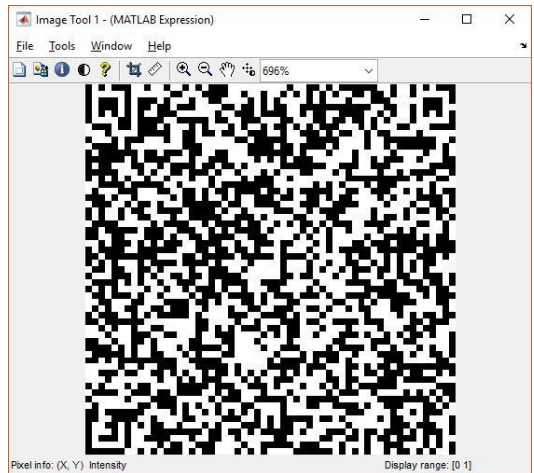

*Figure 3 An example of a gaussian noised QR code*

Z is a standard normal variable. The pixels whose values are flipped are those which have a noise |X|>0.5 added to them. Also, pixels=1 are only flipped if a noise X>0.5 is added to them, while pixels=0 are flipped if X<-0.5 is added to them. Using this and the symmetry of the pdf of a standard normal variable we can calculate the expected value of the number of pixels which are flipped. Now the expected value of the number of pixels which are flipped can be calculated as-

$$E(fraction\ of\ bits\ which\ are\ flipped)$$
$$= 1 - F_z\left(\frac{0.5}{0.54}\right) = 1 - F_z(0.9) = 0.29$$

Where $F_z$ is the Cumulative density function of a standard normal variable.

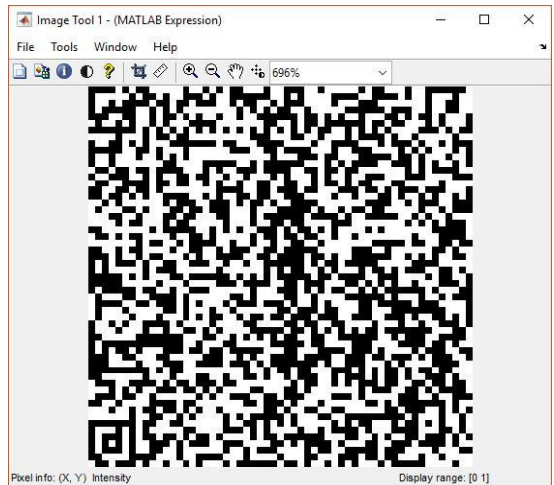

*Figure 4 An example of a QR code having salt and pepper noise*

**b) Salt and pepper noise** – We also tested our network on QR code images having salt and pepper noise with a fraction equaling 0.4 of the total pixels being affected. We used the following function for introducing noise in our images- Imnoise (I,'salt & pepper'd) function of MATLAB [15] where d is the fraction of pixels being affected by the noise. This is a very common type of noise present in binary images.

**c) Parts of the QR code affected by huge amounts of noise-** Another type of noisy QR codes which we tried our network on are QR codes with some part of them being highly corrupted. We corrupted 1 corner of the QR code (around 1200 pixels) with huge amounts of salt and pepper noise (with d=1). Also, we completely blacked out and whited out a part of the QR code (about 1200 pixels). This image was then given as input. This type of degradation of QR codes is common in the real world where a corner of a printed code may get degraded, or a QR code torn off. Also, this type of noise would model the parts of a QR code being scanned in a partial shadow or being blurred by motion.

*Figure 5 left corner having a large amount of noise/blacked out*

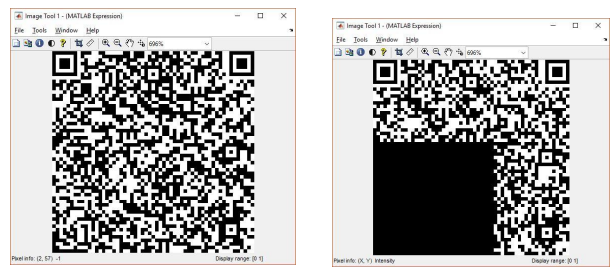

*D. Parallel Algorithm*

Whenever a noisy input is presented to the network, it is first converted to a 3249×1 vector. The image-

$$I = \begin{bmatrix} x_{1,1} & x_{1,2} & \cdots & \cdots & x_{1,56} & x_{1,57} \\ \vdots & & & & \vdots & \end{bmatrix}_{57 \times 57}$$

This is converted to a 3249×1 vector-

$$V = \begin{matrix} x_{1,1} \\ \vdots \\ x_{57,57} \end{matrix}$$

And then we convert it to a bipolar vector {1, -1} form from the {0,1} form a usual binary image is in, by multiplying it by 2 and subtracting 1 from each entry-

$$S = (2 \times V) - \begin{matrix} 1 \\ \vdots \end{matrix}$$

Then, we calculated the energy of this noisy input in relation with each of the Hopfield network's (and their weight matrices) using the formula-

$$E_k = -\frac{1}{2} \sum_{i,j} W_{k,ij} \times s_i \times s_j$$

Where $W_{k,ij}$ represents the $(I,j)^{th}$ element of the $k^{th}$ networks weight matrix. The $s_i$ and $s_j$ represent the $i^{th}$ and the $j^{th}$ element/node of the Hopfield network (their value) which is equal to the corresponding element of the noisy input vector's (S) value.

This energy of a state in a Hopfield network is a measure of how 'stable' a state of the network is. It indicates how close to convergence a network is. The trained networks energy functions have a minima at the points which are the trained patterns in that network. Also, each iteration of the update of the network involves updating one node (asynchronously) at a time using the rule-

$$s_i = sgn\left(\sum_j W_{i,j} \times s_j\right)$$

sgn is the signum function

Each such iteration leads to a decrease in energy (or at least its energy remaining the same) of the network, as a node is updated to be of the same sign as the above summation, leading to a negative contribution to the overall energy of the network [16]. A Hopfield network converges by eventually settling to a minima in its energy function. Trained patterns also correspond to energy minima's in the network and are hence stable points.

Then, for selecting which network this noisy input belongs to (it belongs to only one of these Hopfield networks as each QR code was only trained in 1 network), we run all the networks with the same noisy input for a small number of iterations in comparison to the number of iterations the network takes for convergence to a stable minima(say 100<<30,000, the usual convergence iterations for our QR code network).

Let the state of all these networks after running for that pre-decided number of iterations be S'$_k$. Now we calculate the energy of these states S'$_k$ using the formula-

$$E'_k = -\frac{1}{2} \sum_{i,j} W_{k,ij} \times s'_{k,i} \times s'_{k,j}$$

The running of these networks for a few iterations would have caused the energy of each of the networks to decrease. This is because with each iteration, for any weight matrix the energy of the Hopfield network either decreases or remains the same [16]. The Hopfield networks running in parallel all converge towards their nearest local minima, which are spurious minima for all of the networks except for one, the one in which the denoised version of the QR code is stored. This is because there are a large number of spurious minima in comparison to the actual trained minimas (QR codes) in a network [17]. So, the probability that the network in which the denoised version of this QR code is not stored will converge to a spurious minima is very high. The one in which the denoised version is trained will have the denoised version of the QR code as a minima very near (in hamming distance) to the noisy input, and hence the noisy code would converge to the denoised QR code in this one network. So, now to get the network in which our denoised QR code is stored, we have to differentiate between the network's converging to a trained minima and network's converging to spurious minima. This can be done using the difference in the energy profile of spurious and trained minimas [18]. Spurious minima have a much shallower (lower change with iterations) well around them than actual trained minimas. So, if we calculate delta-

$$\delta_k = E_k - E'_k$$

Then the networks converging to a spurious minima would have a much lower $\delta_k$ than the one network converging to the trained (denoised QR code) minima. This is because the energy of the network converging to the trained minima would have very rapidly decreased in a few steps. So, the network with the highest $\delta_k$ is selected to be the correct network and only this network is then further run till it converges. The output of this network is the denoised version of the QR code.

## IV. RESULTS

We tested our network for with various types of noise including gaussian and salt and pepper noise.

For gaussian noise, the network had denoised the codes successfully, demonstrating its effectiveness on a fairly common kind of noise.

Such QR codes with gaussian noise of variance=0.3 were completely denoised by our network, as seen in figure 6,7 and 8. The denoised QR code exactly matches the original one.

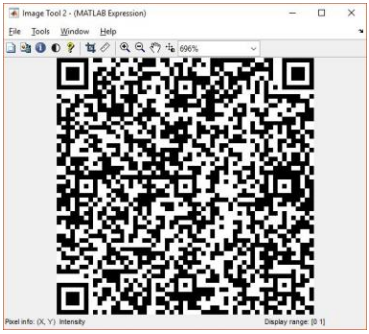
*Figure 6 The original QR code*

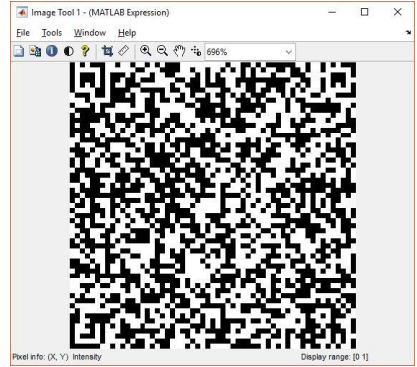
*Figure 7 The noisy QR code with gaussian noise of variance=0.3*

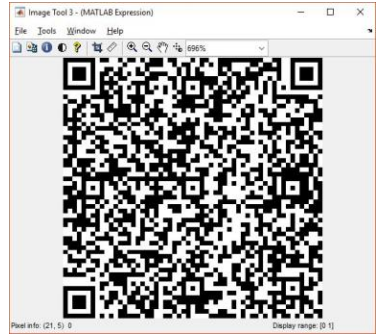
*Figure 8 The denoised QR code*

We also tested denoising QR codes with salt and pepper noise in our network. Such noise is routinely found in binary images due to random thermal fluctuation in cameras- Salt and pepper noise affecting the whole QR code was effectively removed and the original QR code was completely recovered, as seen in figures 9,10,11 .

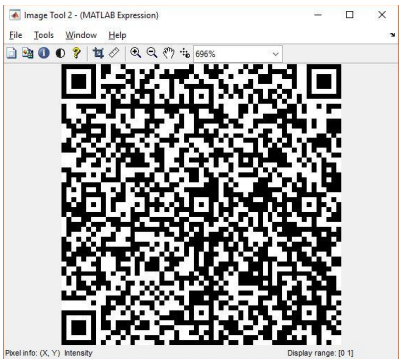
*Figure 9 The original QR code*

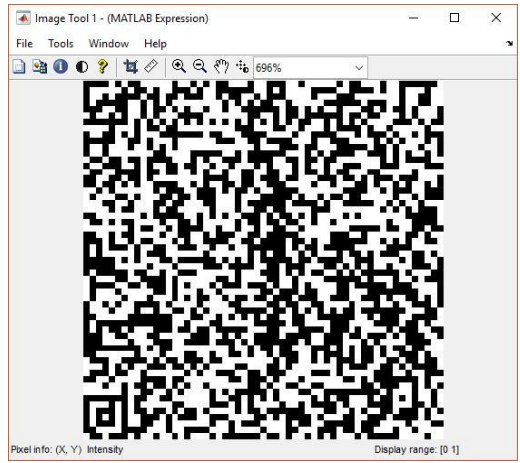
*Figure 10 The noisy QR code with salt and pepper noise throughout the QR code*

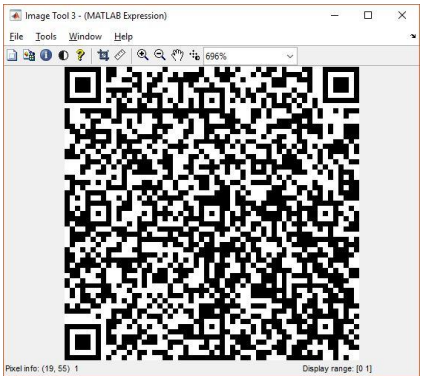
*Figure 11 The denoised QR code*

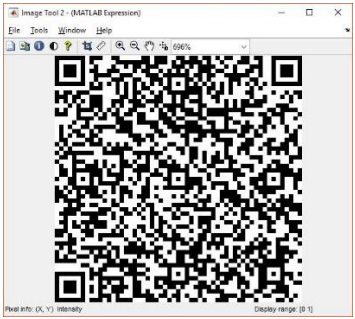
*Figure 12 The original QR code*

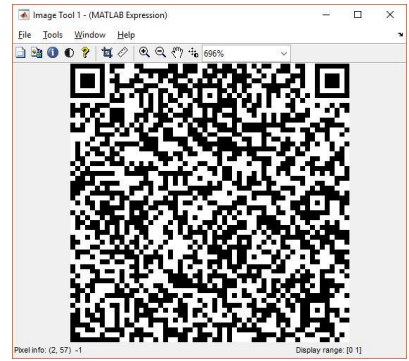
*Figure 13 The noisy QR code with its left corner being affected by a large amount of salt pepper noise*

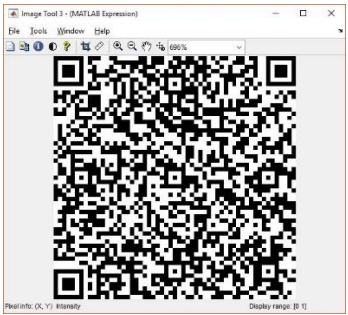
*Figure 14 The denoised QR code*

We also tested our network on QR codes which in which a region is very noisy or completely destroyed. This type of situation does happen in real world scenarios(especially when the QR code is kept outdoors and exposed to wear and tear) and our network should perform well in such situations too.

Such QR codes with a corner affected by high amounts of salt and pepper noise for also completely denoised by our network, as seen in figure 12,13,14.

QR codes with 1 corner being blackened/ whitened out should also be denoised and recovered with our network. This is important as at many times, in outdoor use, parts of the QR code may get destroyed (torn off/ blackened away). They can still be recognized, without requiring any repair/ replacement by our network -

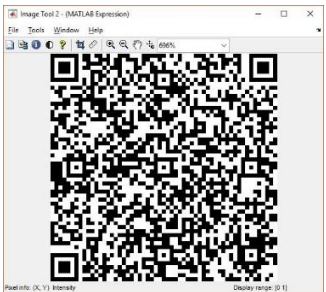
*Figure 15 The original QR code*

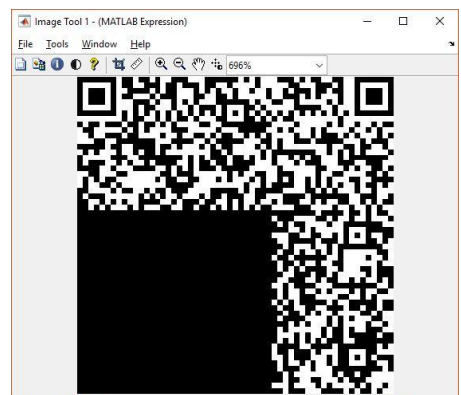
*Figure 16 The noisy QR code with its left corner blacked out*

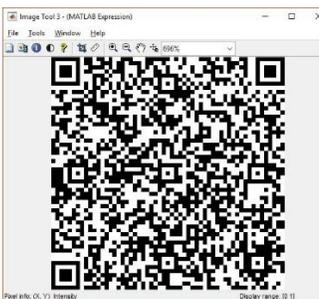
*Figure 17 The denoised QR code*

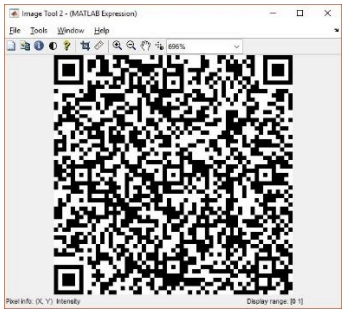
*Figure 18 The original QR code*

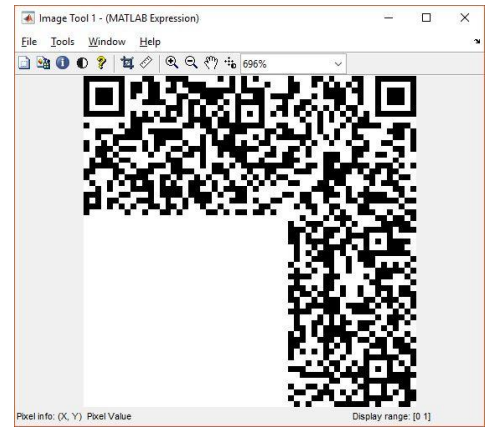
*Figure 19 The noisy QR code with its left corner whited out*

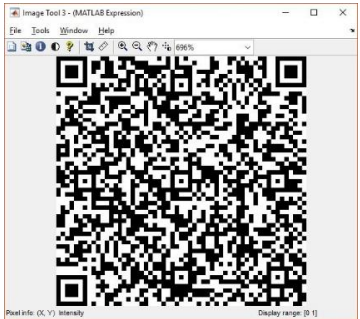
*Figure 20 The denoised QR code*

The Hopfield network accurately denoised QR codes with varying types of noise. It also correctly denoises and recognizes QR codes with high amounts of noise localized in a few regions. The storage capacity of QR codes for a system of n networks in our case was-

$$400 \times n$$

where n is the number of networks used in parallel. Also, it is much faster than a single Hopfield network having the same capacity, as if our network has 3249 nodes and a capacity of $400 \times n$, a single Hopfield network would have $3249 \times n$ nodes as capacity of a Hopfield network scales linearly. Such a network with $3249 \times n$ nodes would be a lot slower to converge, as it has a much larger weight matrix- of the order $(3249 \times n)^2$ terms while each of our network would only be only of the order of $(3249)^2$ terms. This would make the update much quicker. Also, our algorithm would use up less memory than a single Hopfield network of the same storage capacity. A single network with the same storage capacity would require a storage of the order $(3249 \times n)^2$ terms. Our algorithm would require much lesser storage of the order of $n \times (3249)^2$. This means that our algorithm leads to huge gains in speed and storage requirements over usual Hopfield networks, making them feasible for a wider variety of purposes.

## V. CONCLUSIONS

Denoising using Hopfield networks provides a new method for denoising QR codes. Large amounts of noise can be tolerated by our denoising method as seen in the results. Our algorithm for distributing the QR codes in several Hopfield networks and selecting the correct one helps increase the storage capacity of QR codes by the networks thus making it possible to use it in large systems for QR code recognition. It makes use of an abundance of spurious minima's and the energy function of the network to select the network containing the denoised QR code. Also, our method is speedier than storing all the networks in a much larger method as we have multiple smaller networks, which are only run for a very few iteration's (only around 100 in our case). Only one of them (which is also a small network) is run until convergence, thus providing a huge speed benefit. It also reduces the storage cost of the network, decreasing it by n times as compared to a single Hopfield network. This method can be used in various uses of Hopfield network's for increasing the capacity.